\documentclass[letterpaper]{article} 
\usepackage{aaai2026}  
\usepackage{times}  
\usepackage{helvet}  
\usepackage{courier}  
\usepackage[hyphens]{url}  
\usepackage{graphicx} 
\usepackage{natbib}  
\usepackage{caption} 
\usepackage{algorithm}
\usepackage{algorithmic}
\usepackage{amsfonts}
\usepackage{newfloat}
\usepackage{listings}
\DeclareCaptionStyle{ruled}{labelfont=normalfont,labelsep=colon,strut=off} 
\floatstyle{ruled}
\newfloat{listing}{tb}{lst}{}
\floatname{listing}{Listing}
\usepackage{subfigure}
\usepackage{amsmath}
\usepackage{booktabs}
\usepackage{multirow}
\usepackage{colortbl}

\title{Sparse Tuning Enhances Plasticity in PTM-based Continual Learning}

\author{
    Huan Zhang\textsuperscript{\rm 1},
    Shenghua Fan\textsuperscript{\rm 1}, 
    Shuyu Dong\textsuperscript{\rm 2},
    Yujin Zheng\textsuperscript{\rm 1},
    Dingwen Wang\textsuperscript{\rm 1*},
    Fan Lyu\textsuperscript{\rm 3\thanks{
    Corresponding authors: Dingwen Wang and Fan Lyu.}}
}
\affiliations{
    \textsuperscript{\rm 1}School of Computer Science, Wuhan University, Wuhan, China\\
    \textsuperscript{\rm 2}State Key Laboratory of Green Pesticide, Central China Normal University, Wuhan, China\\
    \textsuperscript{\rm 3}New Laboratory of Pattern Recognition, Institute of Automation, Chinese Academy of Sciences, Beijing, China\\

    Emails: \{cszhanghuan, fanshenghua, zhengyujin, wangdw\}@whu.edu.cn; dongshuyu@ccnu.edu.cn;fanlyu@ia.ac.cn \\ 
}

\usepackage{bibentry}

\begin{document}

\maketitle

\begin{abstract}
Continual learning with Pre-trained Models (PTMs) holds great promise for efficient adaptation across sequential tasks. However, most existing approaches freeze PTMs and rely on auxiliary modules like prompts or adapters, limiting model plasticity and leading to suboptimal generalization when facing significant distribution shifts. While full fine-tuning can improve adaptability, it risks disrupting crucial pre-trained knowledge. In this paper, we propose Mutual Information-guided Sparse Tuning (MIST), a plug-and-play method that selectively updates a small subset of PTM parameters, less than 5\%, based on sensitivity to mutual information objectives. MIST enables effective task-specific adaptation while preserving generalization. To further reduce interference, we introduce strong sparsity regularization by randomly dropping gradients during tuning, resulting in fewer than 0.5\% of parameters being updated per step. Applied before standard freeze-based methods, MIST consistently boosts performance across diverse continual learning benchmarks. 
Experiments show that integrating our method into multiple baselines yields significant performance gains.
Our code is available at \url{https://github.com/zhwhu/MIST}.
\end{abstract}

\section{Introduction}
Continual Learning (CL)~\cite{lyu2021multi} is a paradigm in which tasks are learned sequentially, aiming to reduce forgetting of previously acquired knowledge while integrating new information. 
Recently, Pre-Trained Models (PTMs) ~\cite{han2021pre} have shown potential to enhance learning efficiency in CL tasks. By fine-tuning, PTMs can be easily adapted to various downstream tasks, enabling continual learners to acquire new task-specific knowledge more effectively and improving resilience to catastrophic forgetting~\cite{sun2022exploring}.
One important challenge of PTMs in CL lies in how to effectively adapt to incremental tasks without harming the generalization ability of PTMs.

A common practice is to freeze the PTM and introduce additional learnable parameters to adapt the frozen PTM to new tasks. These methods can typically be categorized into two types: prompt-based methods and adapter-based methods. Prompt-based methods, such as L2P~\cite{wang2022l2p} and DualPrompt~\cite{wang2022dualprompt} introduce additional learnable prompt pools, which dynamically guide the frozen pre-trained layers to accommodate incremental tasks.
Adapter-based methods, such as APER~\cite{zhou2025revisiting} and RanPAC~\cite{mcdonnell2023ranpac}, adapt the frozen PTM by introducing additional adapters during the initial incremental stage to bridge the domain gap between pre-trained representations and incremental task distributions.
In summary, most PTM-based CL methods typically freeze PTMs during incremental learning, relying heavily on the pure pre-trained knowledge for downstream adaptation. 
However, new task distributions may deviate from the encoded PTM knowledge, the freezing backbone struggles to generalize effectively across all tasks, that is, poor plasticity~\cite{zhang2023slca}.
Since freezing PTMs can reduce model plasticity, it raises the question of why some methods that fine-tune PTMs still achieve suboptimal performance. 
A possible explanation lies in the fact that heuristic fine-tuning or fully updating all parameters can lead to the loss of crucial parameters, diminishing the effectiveness of PTMs themselves. To avoid this, effective sparse tuning is needed, which selectively updates only a subset of parameters, thus preserving key knowledge within PTMs. 
\textit{The goal of this paper is to propose a sparse update method that strikes a balance between effective adaptation to new tasks and the preservation of generalization in PTM-based CL methods}.

\begin{figure*}[t]
\centering
  \subfigure[Prompt-based CL methods]{
    \includegraphics[width=0.652\textwidth]{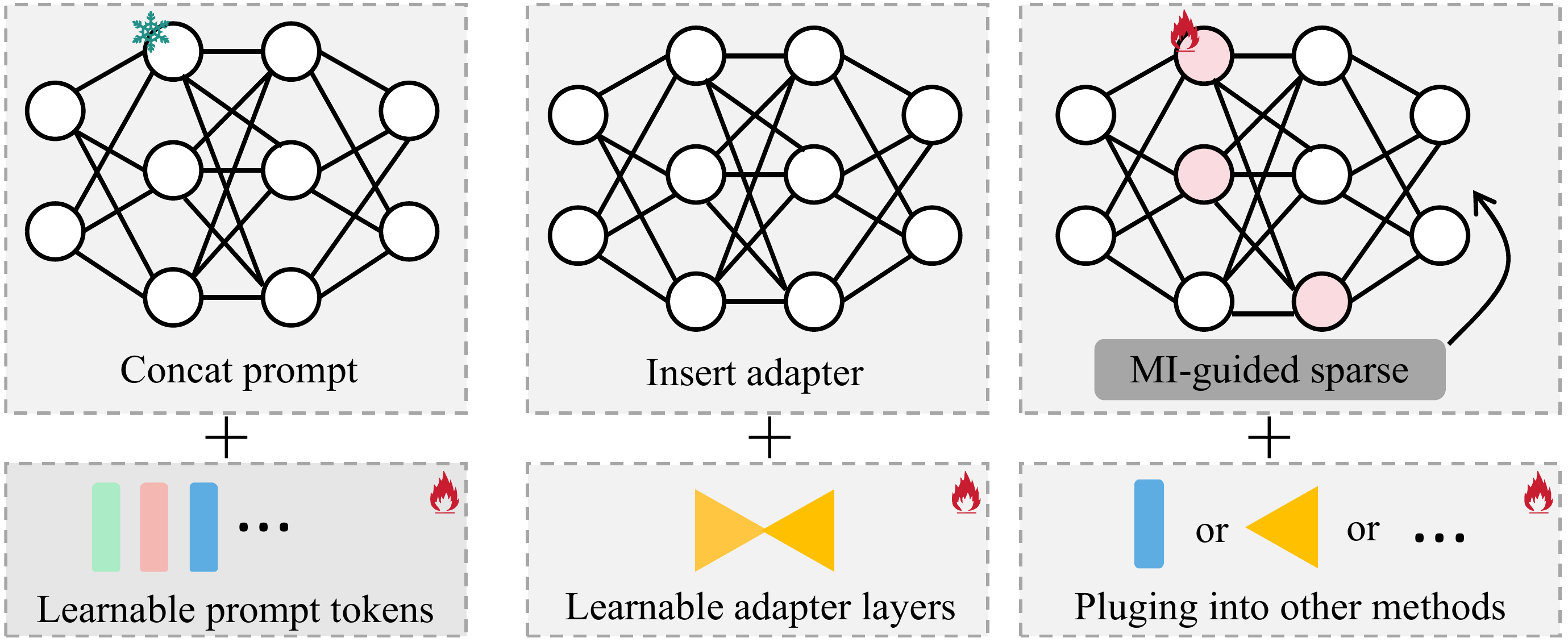}
  }
  \hfill 	
  \subfigure[Plasticity Comparison]{
    \includegraphics[width=0.322\textwidth]{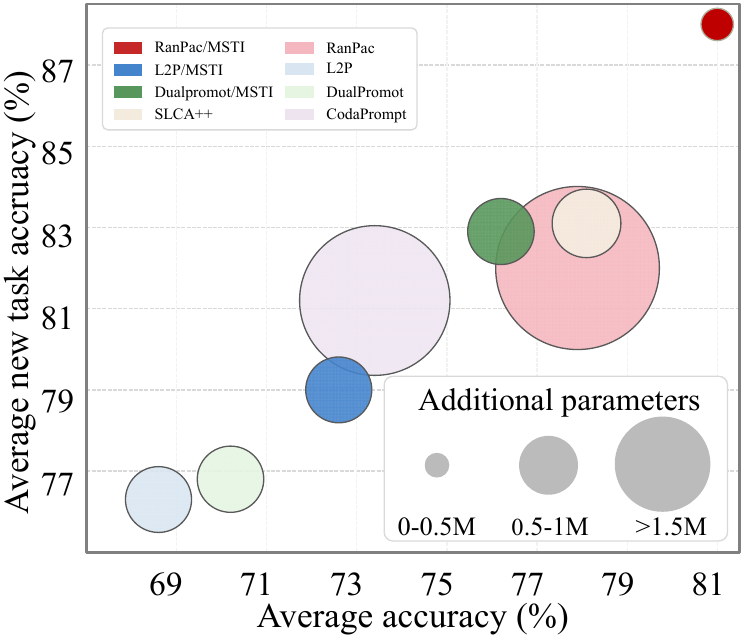}
  }
\caption{
(a) \includegraphics[height=.8em]{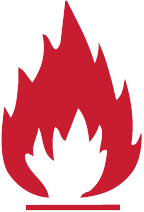} indicates learnable parameters, while \includegraphics[height=.8em]{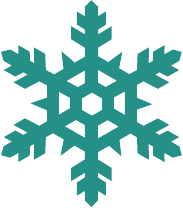} denotes frozen parameters. MIST leverages MI for pre-adaptation, enabling it to be plugged into other methods.
(b) Comparison of different methods in terms of average accuracy, new task accuracy, and additional parameters. 
MIST achieves superior accuracy through MI-guided sparse tuning.
}
\label{fig:fig1}
\end{figure*}

Therefore, how to selectively identify important parameters in PTM-based CL remains a key challenge. To address this, we investigate the underlying behavior of PTMs through a probabilistic analysis. We theoretically and empirically demonstrate that the parameters sensitive to the MI objective can effectively model task-specific knowledge while minimizing disruption to the original knowledge structure of the PTM. Motivated by this, we introduce a simple yet effective plug-and-play method named \textbf{Mutual Information-guided Sparse Tuning} (MIST). Specifically, before training each incremental task with other freeze-based methods, we first determine the sensitivity of each PTM parameter to the MI objective. We then select the top 5\% most sensitive parameters for MI-guided tuning
, which enables the model to fully adapt to the new task distribution while maximally preserving the structural knowledge encoded in the PTM. 
During this process, we apply strong regularization by randomly dropping the gradients of 90\% of the selected parameters in each mini-batch, thereby updating only 0.5\% of the parameters per batch. After this tuning stage, we freeze the PTM and proceed with the original freeze-based method for continual learning. We insert our approach into six representative freeze-based methods and conduct experiments on serval datasets. Results show consistent performance improvements with MIST, particularly on datasets with large distribution shifts from the pretraining domain. For example, SimpleCIL with MIST achieves 17.9\% and 15.7\% improvements on Split-ImageNet-R and Split-Cars, respectively.
The contributions of this paper are summarized as follows:
\\
(1) We study PTM-based CL from a probabilistic and information-theoretic perspective, and theoretically and empirically demonstrate, through MI techniques, PTMs can effectively adapt to new tasks by updating only a small subset of parameters.\\
(2) We introduce a simple yet effective plug-and-play method named {Mutual Information-guided Sparse Tuning} (MIST), which can be integrated into freeze-based methods to provide significant performance improvements.\\ 
(3) We incorporate MIST into six representative PTM-based CL methods and evaluate them across five benchmark datasets. All methods achieve consistent performance gains after integrating MIST, highlighting its broad applicability and effectiveness. The empirical results clearly demonstrate the superiority of MIST.

\section{Related Work}
\textbf{Continual Learning on Pre-trained Models}.
Recently, advancements in PTMs and their exceptional performance in adapting to downstream tasks have inspired researchers to investigate how PTMs can be adapted for continual learning across sequential tasks. 
Prompt-based methods learn continual prompts to provide fixed PTMs with additional instruction. 
DualPrompt \cite{wang2022dualprompt} combines task-shared and task-specific prompts to achieve an effective balance between adaptability and mitigating forgetting, while CODA-Prompt \cite{smith2023coda-prompt} leverages contrastive learning-based prompts to enhance the representation learning of PTMs for improved task adaptation.
HiDe-Prompt\cite{hide-prompt} optimizes hierarchical components by combining task-specific prompts and representation statistics, enhanced with a contrastive regularization strategy.
Adapter-based methods also freeze the PTM and introduce additional lightweight modules for task-specific adaptation.
SLCA++\cite{zhang2024slca++} sequentially fine-tunes low-rank LoRA matrices with a small learning rate to avoid disrupting the pre-trained features.
RanPAC\cite{mcdonnell2023ranpac} adapts the PTM during the first task to enhance downstream performance, while
APER~\cite{zhou2025revisiting} further combines the adapted PTM with the original frozen PTM to jointly extract features, aiming to balance generalization and task-specific learning.

\noindent
\textbf{Mutual Information in Machine Learning}.
With the advancement of deep learning\cite{lyu2024elastic,yin2025beyond,zhang2024constructing,du2024confidence}, mutual information (MI) has become an important tool for capturing both linear and nonlinear dependencies between variables, supporting tasks such as feature selection, clustering, and model optimization~\cite{zhang2023mfsjmi, vinh2009information, tishby2015dlib}. In particular, InfoNCE~\cite{oord2018representation} has emerged as a widely used lower-bound estimator of MI in representation learning. Building on this, recent works have applied InfoNCE-based MI objectives to continual learning. For example, Guo et al.\cite{guo2022ocm} used InfoNCE to measure MI between samples to mitigate catastrophic forgetting, while Li et al.\cite{li2023variational} maximized MI between outputs of current and previous models for knowledge distillation.
In this work, we construct an MI-guided sparse tuning to identify important parameters during incremental fine-tuning in PTM-based CL, enabling more targeted and generalization-preserving updates.

\section{Rethinking PTMs in Continual Learning}
\subsection{Continual Learning with PTMs and the Impact of Freezing PTMs}

\textbf{Preliminaries.}
Given a sequence of tasks with data $\{\mathcal{D}^1, \mathcal{D}^2, \dots, \mathcal{D}^T\}$, where $\mathcal{D}^t = \{(x_i, y_i)\}_{i=1}^{n_t}$ with $n_t$ input pair, sample $x$ and its corresponding label $y$. 
Different tasks are with disjoint label spaces across tasks: $\mathcal{Y}_i \cap \mathcal{Y}_j = \emptyset$ for $i \neq j$. At the training stage $t$, only the current task dataset $\mathcal{D}^t$ is available.
The model is denoted as $f_\theta$, where $\theta$ is parameter.

In PTM-based CL,  $\theta$ is initialized from a PTM trained on a large-scale dataset and is typically frozen during adaptation. The model adapts to new tasks by introducing additional parameters, which can take the form of prompts or adapters, depending on the chosen tuning strategy. 
Despite their structural differences, both methods freeze the backbone and optimize lightweight parameters for efficient adaptation. 
Freezing the PTM parameters $\theta$ limits adaptability to new tasks, shifting the burden to auxiliary modules like prompts or adapters.

\noindent
\textbf{Freezing PTMs in prompt tuning}.
In prompt-based tuning (the left subfigure in Fig. \ref{fig:fig1}(a)), learnable prompts $\phi$ are prepended or injected into the input embedding, yielding output $p(y \mid x; \theta, \phi) = f_{\theta}(P_\phi(x))$.
The gradient of the log-likelihood with respect to \( \phi \) follows the chain rule:
\begin{equation*}
    \frac{\partial \log p(y \mid x; \theta, \phi)}{\partial \phi}
    = \frac{1}{p(y \mid x; \theta, \phi)} \cdot 
    \frac{\partial f_{\theta}(P_\phi(x))}{\partial x} \cdot 
    \frac{\partial P_\phi(x)}{\partial \phi},
\end{equation*}
where $P_\phi(x)$ represents the modified input obtained by injecting the learnable prompt $\phi$ into the feature space of $x$. Since \( \theta \) is frozen, the Jacobian term \( {\partial f_{\theta}}/{\partial x} \) is fixed and reflects the model's sensitivity to input perturbations. When this Jacobian is close to zero in directions that encode task-specific features, the gradient signal received by \( \phi \) is significantly diminished, regardless of its expressive capacity~\cite{qiao2023prompt2, fu2024prompt4, gao2023prompt}. This severely restricts the effectiveness of prompt-based tuning, especially under distribution shifts where new tasks require directions outside the pre-trained manifold.

\noindent
\textbf{Freezing PTMs in adapter-based tuning}.
Adapter-based tuning (the center subfigure in Fig. \ref{fig:fig1}(a)) inserts trainable adapters $\psi$ into the intermediate layers, resulting in $p(y \mid x; \theta, \psi) = f_{\theta, \psi}(x)$.
The gradient of $\psi$ is:
\begin{equation}
    \frac{\partial \log p(y \mid x; \theta, \psi)}{\partial \psi}
    = \frac{1}{p(y \mid x; \theta, \psi)} \cdot \frac{\partial f_{\theta, \psi}(x)}{\partial \psi}
    \label{eq:adapter}
\end{equation}
where $f_{\theta, \psi}$ means the backbone modified by inserting adapter modules. From Eq. \eqref{eq:adapter}, adapter's influence must propagate through the remaining frozen layers to affect the output. If the PTM is not responsive to the features injected by adapters, particularly when such features lie outside the pre-trained distribution, then the resulting gradient with respect to \( \psi \) is similarly attenuated~\cite{qiao2024ada1, son2024ada3, nowak2024ada4}. 

In summary, despite using different mechanisms, both prompt- and adapter-based methods suffer from gradient suppression due to the fixed representational structure of the frozen PTM.
The frozen PTM acts as a bottleneck that limits the flow of gradients to newly introduced parameters. This hampers the model's ability to adapt to novel tasks.

\subsection{Continual Tuning on PTMs}

To enhance plasticity in PTM-based CL, some works have explored direct fine-tuning. However, studies~\cite{adam,zhang2024slca++} show that this often results in significant performance drops, particularly under distribution shifts. To analyze this, we begin by examining the gradient of the log-likelihood:
\begin{equation}
\begin{aligned}
\frac{\partial \log p(y \mid x; \theta)}{\partial \theta^i} = & -\frac{1}{p(x, y; \theta)} \cdot \frac{\partial p(x, y; \theta)}{\partial \theta^i} 
\\ 
&+ \frac{1}{p(x; \theta)} \cdot \frac{\partial p(x; \theta)}{\partial \theta^i},
\end{aligned}
\label{eq:ce_grad}
\end{equation}
where $\theta^i \in \theta$  denotes an arbitrary parameter in $\theta$. In Eq. \eqref{eq:ce_grad} the term ${\partial p(x, y; \theta)}/{\partial \theta^i}$ encourages task-specific alignment through updates to $p(x, y; \theta)$, while the term ${\partial p(x; \theta)}/{\partial \theta^i}$ reflects how parameter changes disturb the pre-trained input distribution $p(x; \theta)$. Excessive increase or decrease in the second term can distort the underlying feature, resulting in poor generalization.

\begin{figure}
    \centering
    \includegraphics[width=0.7\linewidth]{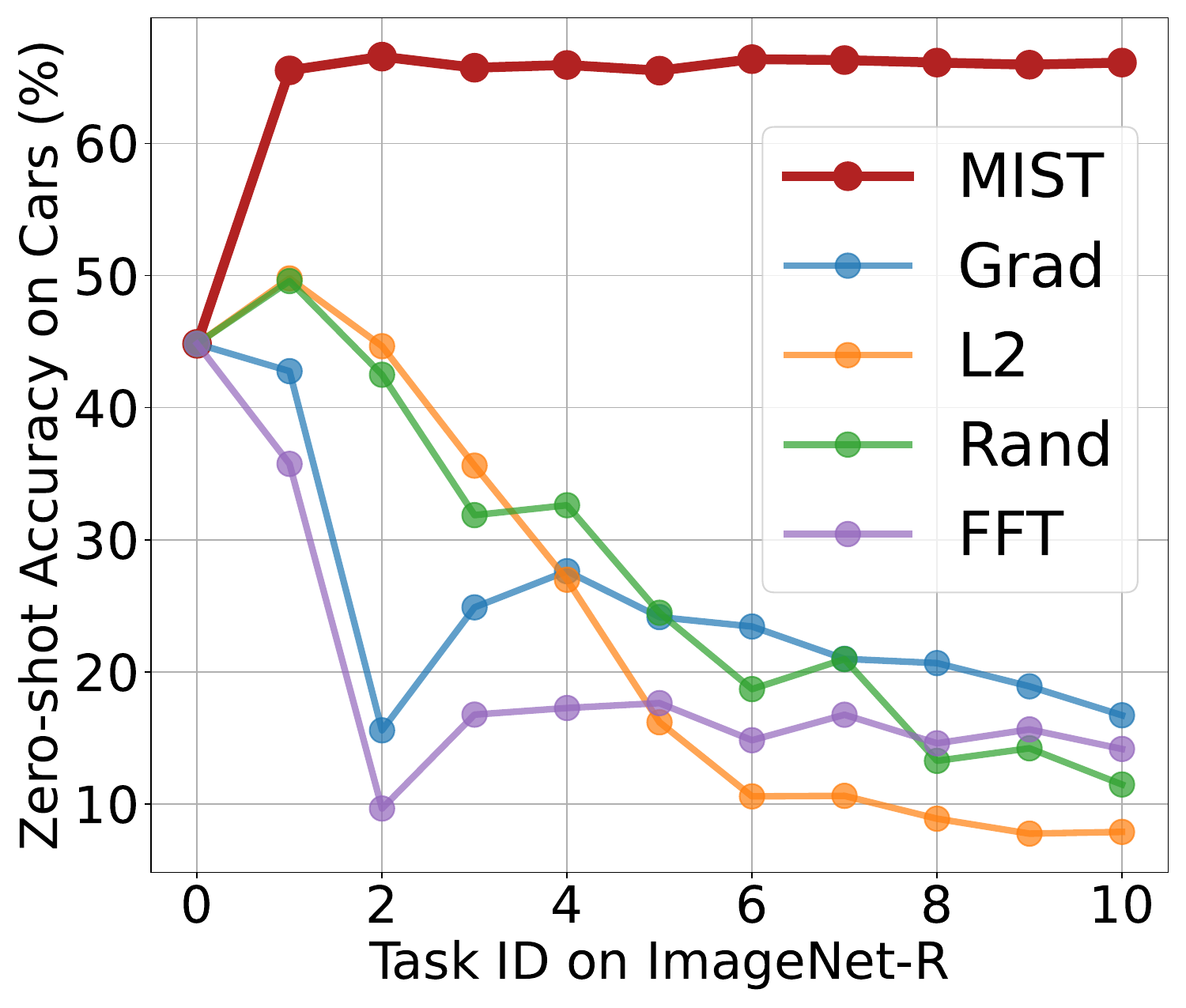}
    \caption{
        Zero-shot accuracy using different sparse update strategies.
        After training on each ImageNet-R task, we freeze the PTM and train only on a new classification head on the full Cars dataset. 
        Higher accuracy reflects better generalization capability of the PTM learned from each task.
    }
    \label{zero_shot}
\end{figure}

Existing tuning strategies, including full fine-tuning, naive partial fine-tuning, and Fisher-guided partial fine-tuning, share a common limitation: they inadvertently perturb the pre-trained structure by amplifying the term ${\partial p(x; \theta)}/{\partial \theta^i}$, thereby severely reducing the generalization of the PTM. As illustrated in Fig.~\ref{zero_shot}, this drawback leads to a continual decline in zero-shot accuracy on the Cars dataset as tasks progress, clearly indicating progressive loss of generalization ability. 
Existing strategies either overfit to new tasks or disrupt pre-trained generalization due to their inability to disentangle task-relevant gradients from those that compromise structural stability. This motivates the need for a more principled tuning strategy that explicitly controls the influence on each gradient component in Eq.~\eqref{eq:ce_grad}. 
This motivates the need for a more principled tuning strategy that explicitly controls the influence on each gradient component.

\section{Method}
\subsection{Mutual Information Analysis in PTM-based CL}

MI~\cite{EstimatingMI, lei2023mutual} is a fundamental concept in information theory and has been widely adopted in machine learning.
By maximizing the MI $I(X;Y)$ between input $X$ and output $Y$, MI explicitly quantifies the statistical dependency between features and labels~\cite{guo2022ocm}. 
Formally, the MI $I(X;Y)$ is defined as:
\begin{equation}
    I(X; Y) = \mathbb{E}_{(x, y) \sim \mathcal{D}^t} \left[ \log \frac{p(x, y; \theta)}{p(x; \theta)p(y; \theta)} \right],
    \label{eq:mutual_info}
\end{equation}
where \( p(y; \theta) \) denotes the prior probabilities of the target classes.
Due to the normalization constraint of probability distributions, we have \( \sum_x p(x; \theta) = 1 \) and \( \sum_y p(y; \theta) = 1 \). Under this constraint, the gradient of the MI with respect to \( \theta^i \) can be simplified as:
\begin{equation}
\frac{\partial I(X; Y)}{\partial \theta^i} 
= \mathbb{E}_{(x, y) \sim \mathcal{D}^t} \left[ \frac{\partial p(x, y; \theta)}{\partial \theta^i} \log \frac{p(x, y; \theta)}{p(x; \theta)p(y; \theta)}\right].
\label{eq:mi_grad}
\end{equation}

In this paper, we explore how MI contributes to the trade-off between plasticity and generalization in PTM-based CL, and make two observations.

\noindent
(1) \textbf{Mutual Information Gradients: Stable Adaptation with Minimal Interference}.
Compared with CE gradients, MI gradients induce less disruption to the pre-trained feature space. Both gradients, as shown in Eq. \eqref{eq:ce_grad} and Eq. \eqref{eq:mi_grad}, include the term $\partial p(x, y; \theta)/\partial \theta^i$, which accounts for task-specific supervision. However, the CE gradient additionally involves the marginal term $\partial p(x; \theta)/\partial \theta^i$, which directly modifies the input distribution learned by the PTM. This term does not appear in MI gradients due to the probabilistic normalization constraint imposed by mutual information objectives, thereby naturally preserving the structural integrity of the input features.
\\
(2) \textbf{Diverse Batches Improve Gradient Stability under MI Objectives}.
The MI gradient formulation assumes a normalization condition $\sum_x p(x; \theta) = 1$, which holds exactly only when the full data distribution is observed. In practice, this assumption is better approximated when batches contain a diverse and representative set of samples. Consequently, using larger and more varied batches helps reduce gradient estimation bias and further mitigates unintended shifts in the pre-trained representation space during adaptation.

In summary, MI provides a more stable optimization objective than CE for CL with PTMs. Unlike CE gradients, which include the marginal term $\partial p(x; \theta)/\partial \theta^i$ and may disrupt the pre-trained input distribution, MI gradients inherently avoid this due to normalization constraints, preserving feature integrity. Additionally, MI benefits from diverse batches, which better approximate the underlying data distribution and reduce gradient bias. Together, these properties enable MI to strike a more effective balance between plasticity and stability during adaptation.
\textit{While MI enables a better plasticity–stability trade-off, directly replacing CE for full fine-tuning may still lead to information loss and high computational cost due to large-scale updates.} To address this, we next introduce a lightweight MI-based method that selectively tunes a small parameter subset and can be flexibly integrated as a plugin into existing PTM-based CL methods, including prompt-based and adapter-based approaches.

\subsection{Mutual Information-guided Sparse Tuning (MIST)-A plug-and-play solution}
In this subsection, we introduce Mutual Information-guided Sparse Tuning (MIST), a plug-and-play pre-adaptation framework compatible with a wide range of PTM-based CL methods, including those based on prompt tuning and adapters.
MIST acts as a pre-adaptation stage that sparsely fine-tunes the PTM before one freeze-based method. 
Specifically, it identifies the top-$k\%$ most MI-sensitive parameters through gradient-based sensitivity analysis, and selectively fine-tunes them using a mutual information objective. This pre-adaptation helps reshape the feature space with minimal interference to the pre-trained structure. 

To efficiently estimate the sensitivity of each parameter $\theta^i \in \theta$ to the MI objective, we adopt the MI-based Fisher Information Matrix~\cite{ewc_0} as an importance measure. While computing exact gradients over the entire task is computationally intensive, the sample distribution within a task is typically uniform in CL, enabling a batch-wise approximation:
\begin{equation}
    F_\mathrm{MI} =\left( \frac{\partial \mathcal{L}_{\mathrm{MI}}^{\mathcal{D}t}}{\partial \theta} \right)^2 \approx F'_\mathrm{MI} = \left( \sum\nolimits_{j=1}^{B_j \leftarrow\mathcal{D}_t} \frac{\partial \mathcal{L}_{\mathrm{MI}}^{B_j}}{\partial \theta} \right)^2,
    \label{eq:mifisher_task}
\end{equation}
where $\mathcal{L}_{\mathrm{MI}}^{\mathcal{D}_t}$ is the MI loss computed over task $\mathcal{D}_t$, and $B_j$ represents the $j$-th mini-batch sampled from $\mathcal{D}_t$. In practice, we identify the top $k\%$ of parameters with the highest $F'_\mathrm{MI}$ values as the most MI-sensitive parameters, denoted by $\mathcal{M}$:
\begin{equation}
    \mathcal{M} = \left\{ \theta^i \in \theta \mid \operatorname{rank}(F'_{\mathrm{MI}}(\theta^i)) \leq \lfloor k\% \cdot |\theta| \rfloor \right\}.
    \label{eq:mi_topk}
\end{equation}
where $\operatorname{rank}(\cdot)$ denotes the descending order index, and $\lfloor \cdot \rfloor$ denotes the floor function. That is, we select the top $k\%$ parameters with the highest MI-based importance scores.
Given that only a small subset of parameters is updated in each batch and that PTMs are typically initialized near an optimal solution~\cite{zhang2023slca, zhou2025revisiting}, we compute $F'_\mathrm{MI}$ once at the beginning of each task and reuse $\mathcal{M}$ throughout the pre-adaptation phase to ensure both efficiency and effectiveness.

\begin{figure}[t]
\begin{algorithm}[H]
\caption{\textbf{MI-guided Sparse Tuning}}
\label{alg:mi1}
\begin{algorithmic}[1]
\REQUIRE Continual tasks $ \{\mathcal{D}_1, \dots, \mathcal{D}_T\}$, pre-trained model $f_{\theta}$, select rate $k\%$, dropout rate $d\%$
\FOR{task $t = 1$ to $T$}
    \STATE Compute the Fisher matrix $F'_\text{MI}$ using Eq.~\eqref{eq:mifisher_task}
    \STATE Generate parameters group $\mathcal{M}$ by selecting top $k\%$ parameters using Eq.~\eqref{eq:mi_topk}
    \FOR{each training mini-batch iteration}
        \STATE Compute MI loss using Eq.~\eqref{eq:miloss}
        \STATE Generate dropped  parameters group $\mathcal{M’}$ by dropping $d\%$ parameters in $\mathcal{M}$
        \STATE Update the parameters in $\mathcal{M’}$
    \ENDFOR
    \STATE Training other method's additional parameters (e.g. prompt, adapter and classifer) on $\mathcal{D}_t$
\ENDFOR
\end{algorithmic}
\end{algorithm}
\end{figure}

With the MI-sensitive parameter subset $\mathcal{M}$ identified, we proceed to the pre-adaptation stage using an MI-based objective to minimize disruption to the pre-trained feature structure. However, computing the exact MI loss is challenging in practice, as both joint and marginal distributions $p(x, y; \theta)$ and $p(x; \theta)$ are typically intractable. Inspired by OCM~\cite{guo2022ocm}, we adopt the supervised InfoNCE loss to construct the MI objective:
\begin{align}
    \mathcal{L}_{MI} = \sum_{i=1}^{|B|} \frac{A_i}{3|B| \sum_{s=1}^{|B|} \mathbf{1}(y_{s} = y_{i})},
    \label{eq:miloss}
\end{align}
where $X,Y \in \{x_i,y_i\}_{i=1}^{|B|}$. And $A_i$ is given by:
\begin{align*}
    A_i = -\sum_{y_{k} = y_{i}} \log \frac{
        g(x_i, x_k) \cdot g(x_i, x_k') \cdot g(x_i', x_k)
    }{
        \left( \sum_{j=1}^{|B|} g(x_i, x_j) + g(x_i, x_j') + g(x_i', x_j) \right)^3
    },
\end{align*}
where $g(x_i, x_j') = e^{{f_{\theta}(x_i)^T f_{\theta}(x_j')}/{\tau}}$ is the similarity of two samples, $\tau$ is temperature, $x_j'$ is an augmentation view of sample $x_j$. By optimizing Eq.~\eqref{eq:miloss}, we effectively maximize the MI I(X;Y), thereby modeling $p(x, y; \theta)$ in a task-discriminative manner.

To further reduce the number of parameters being updated, we introduce a lightweight regularization strategy called Gradient Dropout. During each batch of the pre-adaptation stage, we randomly drop $d\%$ of the MI-sensitive parameters in $\mathcal{M}$, resulting in only $k\% \times d\%$ of total parameters being updated per batch. In practice, we set $k\%=5\%$ and $d\%=90\%$, yielding updates to merely $0.5\%$ of all parameters per batch.
This stochastic suppression addresses a critical issue, i.e., repeatedly updating a fixed subset of parameters can constrain the model’s exploration of the optimization landscape, leading to biased shifts in the feature space. By introducing randomness into the gradient flow, Gradient Dropout promotes more diverse and balanced parameter updates, reduces co-adaptation, and further stabilizes the pre-trained representation by mitigating local bias and limiting excessive perturbations.

\subsection{Plugging MIST into PTM-based Continual Learning: The Algorithm}

As shown in Algorithm~\ref{alg:mi1}, we begin by temporarily unfreezing the PTM $f_\theta$ and estimating the sensitivity of each parameter with respect to the MI objective. Based on this, we select the top $k\%$ most sensitive parameters to form the update set $\mathcal{M}$. During the MI-guided tuning phase, we apply Gradient Dropout.
After a few epochs of such sparsified adaptation, the PTM is refrozen, and the standard freeze-based CL procedure resumes.
This pre-adaptation phase introduces small computational overhead and is compatible with a wide range of PTM-based CL methods. For prompt-based approaches, MIST is applied to the PTM prior to prompt tuning. For adapter-based methods, we do not modify the PTM or perform any initial task-specific fine-tuning. Instead, MIST is used as a lightweight pre-adaptation step, after which classifier training proceeds as originally designed.

\section{Experiment}

\subsection{Experimental Setups}
\textbf{Benchmark}.
We consider five representative benchmark datasets and randomly split each of them into 10 disjoint tasks. Including CIFAR-100~\cite{krizhevsky2009cifar}, ImageNet-R~\cite{hendrycks2021many}, ImageNet-A~\cite{hendrycks2021natural}, CUB-200~\cite{wah2011cub} and Cars-196~\cite{krause2013cars}.
Performance is evaluated using the standard CL metric, \textit{Average Accuracy}~\cite{shim2021ASER}, defined as:
$A_t = \frac{1}{t} \sum_{i=1}^{t} R_{t,i}$, where $R_{t,i}$ denotes the classification accuracy on the $i$-th task after training on the $t$-th task.
We report both $A_T$ and $\bar{A}$ in the main paper. Here, $\bar{A}$ denotes the mean of $A_t$ over all tasks: $\bar{A} = \frac{1}{T} \sum_{t=1}^{T} A_t$. It reflects the average accuracy of all classes seen so far after each incremental task.

\noindent
\textbf{Implementation}.
Following previous works~\cite{wang2022l2p, wang2022dualprompt}, we adopt a pre-trained ViT-B/16 backbone~\cite{dosovitskiy2020vit} for all baselines. Except for OCM~\cite{guo2022ocm}, no methods use a buffer; OCM is implemented in an offline setting with a buffer size of 1000. We follow the original implementations by employing the Adam optimizer for L2P, DualPrompt, and CoDA-Prompt, and the SGD optimizer for all other baselines. 
Our method, MIST, is inserted as a plug-in module before each selected baseline and is trained for 20 epochs using the SGD optimizer with a learning rate of 0.0001. In the MI-based selection stage, we select the top $k\% = 5\%$ most sensitive parameters. For each mini-batch, we further apply a dropout rate of $d\% = 90\%$ to the selected parameters, resulting in only 0.5\% of total parameters being updated per batch. MIST solely optimizes the MI loss (Eq.~\ref{eq:miloss}), with the temperature $\tau$ set to 0.5. For each task, MIST first conducts this sparse fine-tuning, after which the corresponding baseline resumes training using its original configuration. We adopt this setting consistently across all datasets in our experiments.

\begin{table*}[t]
\centering
\caption{Performance comparison on various datasets.}
\setlength{\tabcolsep}{8pt}
{
\begin{tabular}{l|cccccccccc}
\toprule
\multirow{2}{*}{\textbf{Method}}  & \multicolumn{2}{c}{\textbf{CIFAR100}} & \multicolumn{2}{c}{\textbf{ImageNet-R}} & \multicolumn{2}{c}{\textbf{ImageNet-A}} & \multicolumn{2}{c}{\textbf{CUB200}} & \multicolumn{2}{c}{\textbf{Cars196}} \\ 
 & $\bar{A}$ & $A_T$ & $\bar{A}$ & $A_T$ & $\bar{A}$ & $A_T$ & $\bar{A}$ & $A_T$ & $\bar{A}$ & $A_T$ \\
 \midrule
EWC~\cite{ewc_0} & 53.2 & 41.2 & 42.3 & 26.2 & 22.0 & 7.9 &57.6 & 37.8 & 46.7 & 24.3 \\
OCM~\cite{guo2022ocm} & 63.6 & 37.0 & 77.0 & 66.7 & 58.1 & 44.4 &85.1 & 71.7 & 65.3 & 50.5 \\
CODA-Prompt~\cite{smith2023coda-prompt} & 91.3 & 86.9 & 78.5 & 73.4 & 63.9 & 52.7 &84.1 & 79.3 & 52.1 & 45.4 \\
SLCA++~\cite{zhang2024slca++} & 94.1 & 91.5 & 83.0 & 77.5 &67.1 &58.7 & 91.0 & 86.7  & 79.2 & 73.8\\
APER(Adapter)~\cite{zhou2025revisiting}& 83.9 & 85.9 & 74.2 & 66.9 & 62.4 & 52.1 &90.5 & 85.6  & 52.8 & 40.5 \\
\midrule
L2P~\cite{wang2022l2p} & 86.7 & 83.3 & 74.5 & 68.6 & 53.9 & 44.9 & 81.7 & 67.4 &53.9 & 39.6 \\ 
\rowcolor{blue!10}
~~+MIST & 89.1 & 86.1 & 77.5 & 72.6 & 56.9 & 51.2  & 82.3 & 71.8& 63.4 & 52.7 \\ 
DualPrompt~\cite{wang2022dualprompt}& 87.4 & 84.0 & 75.2 & 70.2& 55.7 & 47.7 & 82.3 & 68.8 & 53.2 & 41.6 \\ 
\rowcolor{blue!10}
~~+MIST& 89.0 & 86.2 & 80.1 & 76.2 & 60.1 & 53.3 & 83.1 &70.2 & 62.4 & 52.8 \\ 
SLCA~\cite{zhang2023slca} & 94.1 & 91.5 & 81.7 & 77.0 & 67.9& 59.3& 90.9 & 84.7  & 76.9 & 67.7\\
\rowcolor{blue!10}
~~+MIST & 94.8 & 92.2 & 83.6 & 80.0 & 69.9& 61.0& 92.0 & 87.3  & 80.7 & 74.6\\
EASE~\cite{ease}& 91.8 & 87.4 & 81.1 & 75.1& 65.3 & 54.9 &90.9 & 85.8  &38.4 & 27.3 \\ 
\rowcolor{blue!10}
~~+MIST& 92.2 & 88.2 & 81.8 & 76.3 & 66.0& 56.9  & 91.8 & 87.0& 57.0 & 47.2\\ 
SimpleCIL~\cite{zhou2025revisiting}& 87.1 & 81.3 & 61.1 & 54.3& 59.8 & 48.5 &90.9 & 85.6  &38.8 & 27.8 \\ 
\rowcolor{blue!10}
~~+MIST& 87.9 & 82.1 & 79.5 & 72.2 & 65.5& 55.3  & 91.6 & 86.8& 57.0 & 43.5\\ 
RanPAC~\cite{mcdonnell2023ranpac} & 94.0 & 90.8 & 83.2 & 77.9 & 70.1 & 61.4& 92.6 & 88.9  & 82.8 & 74.6 \\ 

\rowcolor{blue!10}
~~+MIST & {95.3} & {92.4} & {84.9} & {81.0} & {72.5} & {62.5}& {93.6} & {90.4 } & {83.0} & {76.4} \\ 
\bottomrule
\end{tabular}}
\label{tab:performance_comparison}
\end{table*}

\subsection{Experimental Results}

\textbf{Overall performance}.
To assess the versatility of MIST, we plug it into six representative freeze-based methods: L2P, DualPrompt, SLCA, EASE, RanPAC, and SimpleCIL. Among them, L2P and DualPrompt are prompt-based methods that freeze the PTM and learn token-like prompts for adaptation. SimpleCIL does not involve any parameter tuning and directly trains a prototype classifier on frozen representations. RanPAC and EASE are adapter-based methods that inserts and fine-tunes lightweight modules in the PTM. SLCA adopts a full fine-tuning strategy with a reduced learning rate.
As shown in Table~\ref{tab:performance_comparison}, incorporating MIST consistently improves all methods across all datasets. For example, DualPrompt/MIST achieves accuracy gains of +1.6\%, +6.0\%, +4.4\%, +1.4\%, and +11.2\% on the five datasets, respectively. 
Furthermore, we observe that many methods perform poorly on the Cars196 dataset. For instance, the $A_T$ of L2P and SimpleCIL are only 39.6\% and 27.8\%, respectively. This is mainly because pretraining knowledge offers limited utility for complex fine-grained vehicle classification, making it particularly challenging for models to adapt to such new domains. After inserting MIST, the final accuracies of L2P and SimpleCIL increase by +13.1\% and +15.7\% respectively, indicating that MIST effectively enhances the model's ability to align with domain-specific structures before classifier training.
Among all methods, RanPAC/MIST achieves the best overall performance, indicating that even well-designed adapter-based methods benefit from the MI-guided tuning stage. 

\begin{figure}[t]
  \centering
  \begin{minipage}[t]{0.499\textwidth}
    \centering
    \subfigure[Cars-196]{%
      \includegraphics[width=0.49\linewidth]{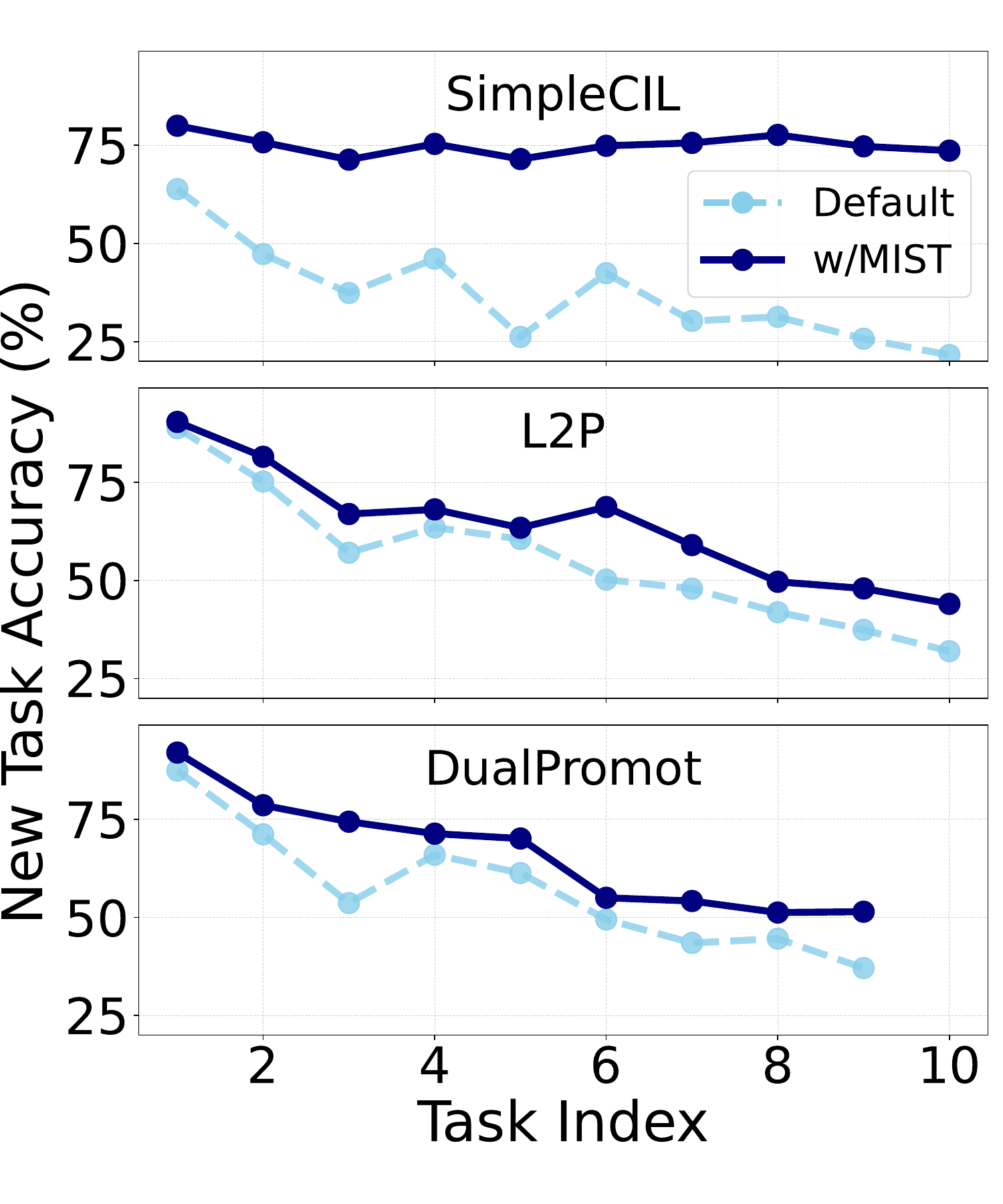}%
    }\hfill
    \subfigure[Imagenet-R]{%
      \includegraphics[width=0.49\linewidth]{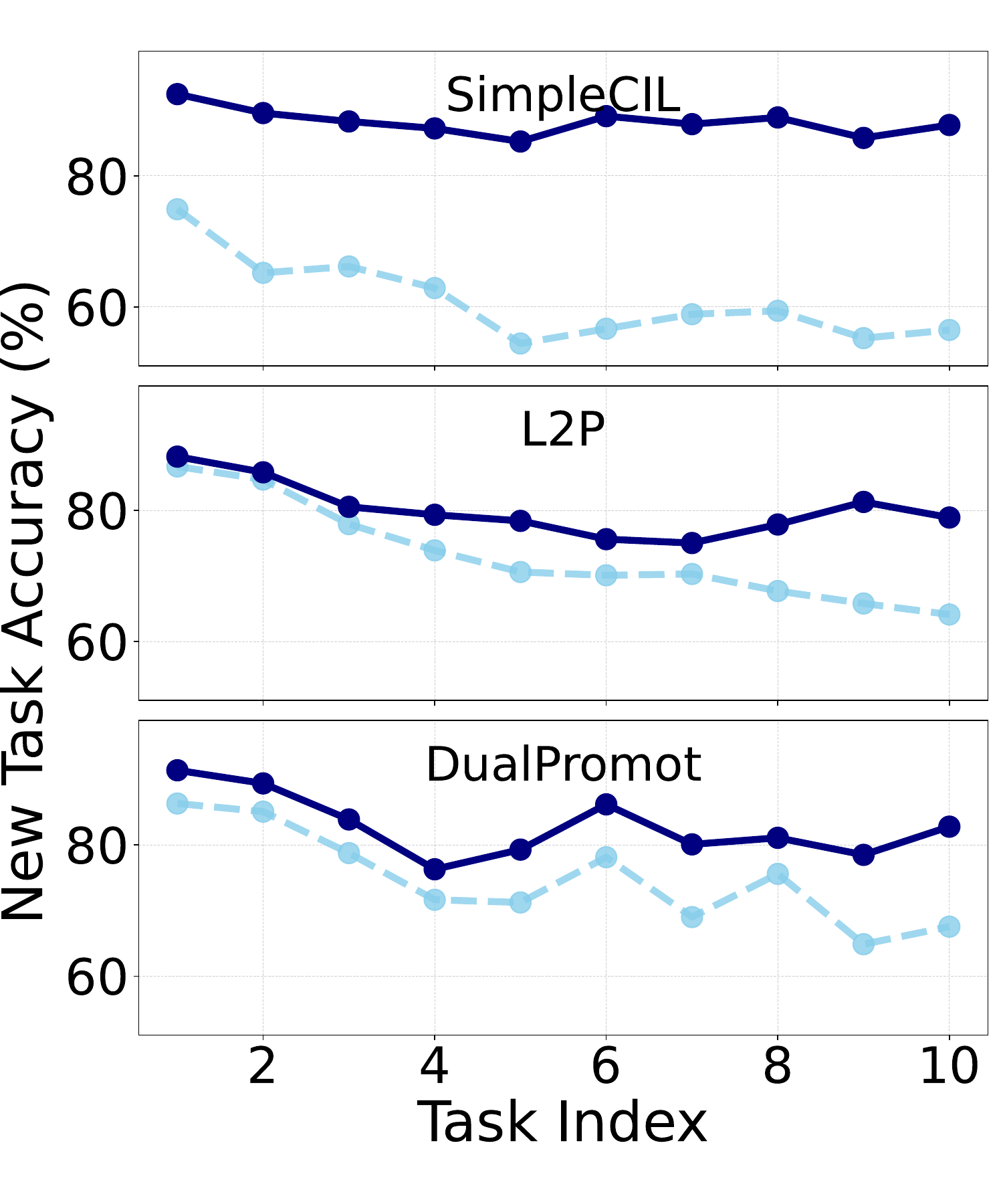}%
    }
    \caption{New task accuracy.}
    \label{fig:new_task_curve}
  \end{minipage}%
  \hfill
  \begin{minipage}[t]{0.49\textwidth}
    \centering
    \subfigure[Cars-196]{%
      \includegraphics[width=0.49\linewidth]{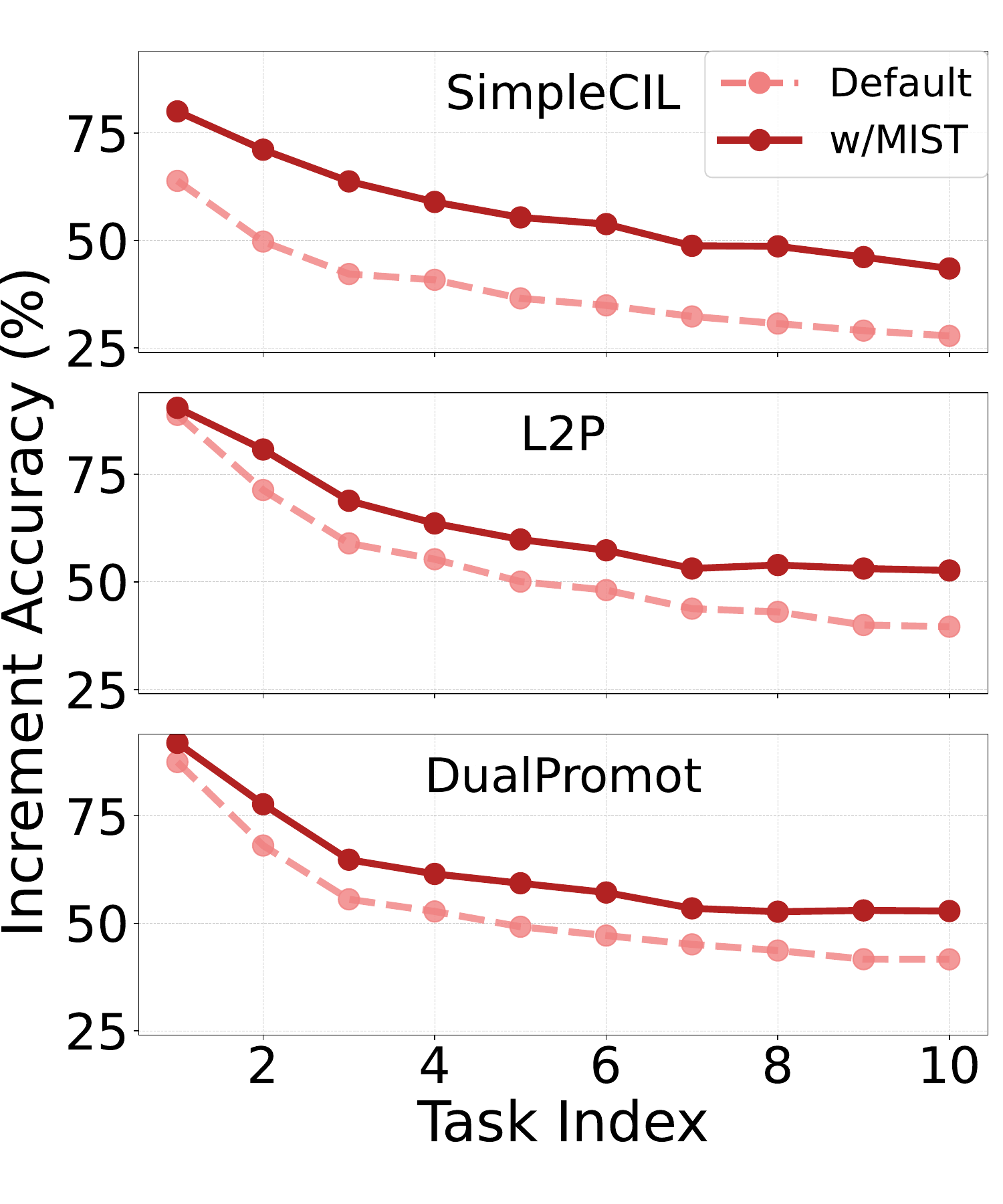}%
    }\hfill
    \subfigure[Imagenet-R]{%
      \includegraphics[width=0.49\linewidth]{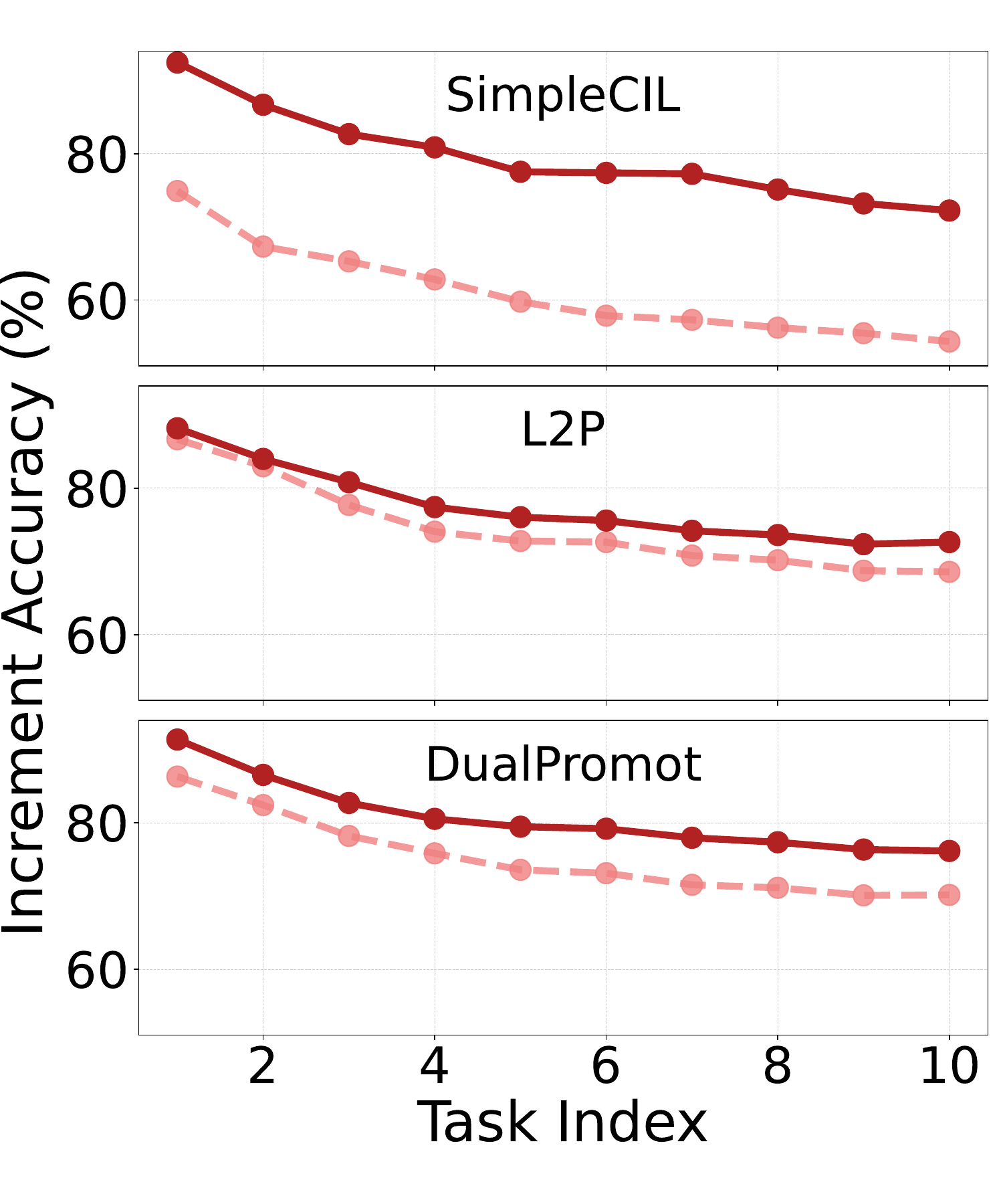}%
    }
    \caption{Incremental accuracy.}
    \label{fig:incremental_curve}
  \end{minipage}
\end{figure}

\noindent
\textbf{Effect of MIST}.
To better understand the effect of MIST on PTM-based CL performance, we visualize both the new task accuracy (Figure~\ref{fig:new_task_curve}) and the incremental accuracy (Figure~\ref{fig:incremental_curve}). New task accuracy refers to the accuracy on the newly learned task, while incremental accuracy denotes the average accuracy over all tasks learned so far. As shown, MIST improves the learning effectiveness across all inserted methods. Specifically, in Figure~\ref{fig:new_task_curve}, the new task accuracy increases significantly for all methods after integrating MIST. This improvement indicates that the pre-adaptation phase provided by MIST helps the PTM align more effectively with task-specific distributions.
Correspondingly, Figure~\ref{fig:incremental_curve} shows that MIST also leads to notable gains in incremental accuracy across all tasks. This is attributed to the improved learning efficiency on new tasks, which in turn contributes to higher cumulative accuracy when evaluated on all seen classes. Taken together, these results highlight the effectiveness of MIST as a plug-in component that improves the task-specific learning capacity of freeze-based methods while preserving their stability, ultimately leading to consistent performance gains in CL scenarios.

\begin{table}[t]
\centering
\caption{Comparison of tuning strategies used for pre-adaptation in RanPAC. “MS” refers to MI-based sparse selection, “ML” to MI loss, and “Drop” to dropout. “MS+ML+Drop” corresponds to our MIST implementation. (a) shows naive selection strategies, (b) adds MI loss and dropout to a naive strategy, and (c) uses MI-based sparse selection combined with MI loss and dropout.}
\setlength{\tabcolsep}{6pt}
{
\begin{tabular}{c|l|cccc}
\toprule
 & \textbf{Method}  & \multicolumn{2}{c}{\textbf{ImageNet-R}} & \multicolumn{2}{c}{\textbf{Cars196}} \\ 
 &  & $\bar{A}$ & $A_T$ & $\bar{A}$ & $A_T$ \\
\midrule
 & RanPAC & 83.2 & 77.9 & 82.8 & 74.6 \\
\midrule
 & ~~+FFT & 57.6 & 36.7  & 31.3 & 11.2 \\
(a)    & ~~+Rand & 52.4 & 11.8 & 33.7 & 10.9 \\ 
    & ~~+L2  & 39.0 & 14.4 & 29.6 & 10.4\\ 
\midrule
 & ~~+Grad & 54.8 & 33.9 & 29.9 & 10.6\\ 
   (b) & ~~+Grad+ML & 76.7 & 67.1 & 64.2 & 40.3 \\ 
    & ~~+Grad+ML+Drop & 77.5 & 68.3 & 64.9 & 42.0 \\ 
\midrule
 & ~~+MS & 75.5 & 66.8 & 76.0 & 66.2 \\ 
    (c)& ~~+MS+ML & 82.2 & 76.7 & 81.3 & 72.1 \\ 
\rowcolor{blue!10}
    & ~~+MS+ML+Drop & \textbf{84.9} & \textbf{81.0} & \textbf{83.0} & \textbf{76.4} \\ 
\bottomrule
\end{tabular}}
\label{tab:compare_tune}
\end{table}
\noindent
\textbf{Comparison with naive sparse tuning strategies}.
We integrate several pre-adaptation strategies into RanPAC, including full fine-tuning (FFT) of all parameters, top 5\% selection based on gradient magnitude (Grad) or parameter norm (L2), and random 5\% selection (Rand). As shown in Table~\ref{tab:compare_tune}(a) and (b), all these alternatives perform significantly worse than MIST.
This is because MIST leverages mutual information (MI) to assess parameter sensitivity, enabling effective task adaptation while minimizing disruptions to the pre-trained representations. In contrast, the alternative methods ignore the importance of preserving pre-trained knowledge during parameter selection, making them vulnerable to catastrophic forgetting and performance collapse. Overall, MIST offers a more balanced adaptation path by jointly preserving plasticity and stability, demonstrating its superiority as a general plug-in pre-adaptation module.

\noindent
\textbf{Ablation study}.
Table~\ref{tab:compare_tune}(c) presents the ablation study of MIST. When applying only MI-based sparse selection ("MS"), the model achieves the lowest accuracy among the three. Although sparse selection reduces parameter interference, the optimization remains guided by the cross-entropy loss, which fails to explicitly preserve the pre-trained feature distribution. Nevertheless, this approach still outperforms alternative selection strategies. Upon introducing the MI loss ("MS+ML"), performance improves, indicating that the MI objective effectively guides the model toward downstream distributions while retaining generalization. Finally, adding gradient dropout ("MS+ML+Drop") yields the highest accuracy across all settings, as it regularizes the update path and mitigates overfitting to static parameter importance. To further validate the effectiveness of MI-based sparse selection, we also apply MI loss and dropout to the Grad-based strategy in Table~\ref{tab:compare_tune}(b). Although this leads to some performance gains, its results remain clearly inferior to MIST. This highlights the unique advantage of MI-guided selection and the complementary role of the three components in MIST. These findings confirm that MI sparsity, MI loss, and dropout contribute synergistically to overall performance gains.

\begin{table}[t]
    \centering
        \caption{
          Efficiency analysis. 
        }
        \label{tab:effiency}
        \begin{tabular}{lcccc}
        \toprule
        Method & $\Delta$ P (M) & FLOPs (M) & Time (ms) \\
        \midrule

        SLCA &  85.40 & 171.6& 12.3 \\
        RanPac& 4.53 & 8.6& 9.1\\
        L2P & 0.48 & 1.0& 12.4 \\
        MIST& 0.43 & 0.8& 9.7 \\
        \bottomrule
    \end{tabular}
\end{table}

\noindent
\textbf{Parameter efficiency analysis}. 
Table~\ref{tab:effiency} compares the efficiency of different methods in terms of (1) $\Delta$P: the number of parameters updated per mini-batch (in millions), (2) FLOPs: flops for updating the selected parameters per batch, and (3) Time: time required to train a batch on an NVIDIA RTX 4090 GPU.
Among the methods, SLCA performs full fine-tuning and thus has the highest update cost—both in terms of parameters (85.40M) and time (12.3ms). L2P only updates prompt tokens, but incurs additional overhead (12.4ms) likely due to its key-query matching mechanism during token routing. 
MIST, while updating only 0.43M parameters per task, incurs slightly higher computation time (9.7ms) compared to RanPAC. This is because computing the MI loss requires augmented views of each sample. In summary, MIST achieves the lowest update cost without introducing any additional parameters, making it easily pluggable into other methods. 

\section{Conclusion}
In this paper, we investigate the fundamental challenge of balancing plasticity and generalization in PTM-based CL. We reveal that direct fine-tuning often compromises the pre-trained feature distribution, while existing freeze-based methods suffer from limited adaptability to new tasks. Through a theoretical lens grounded in MI, we analyze how gradients derived from MI objectives offer a more stable optimization path by avoiding unnecessary perturbations to the PTM.
Motivated by this, we propose Mutual Information-guided Sparse Tuning, a lightweight and plug-and-play pre-adaptation strategy that selectively updates only the most informative parameters before each incremental task. By computing an MI-based Fisher Information Matrix, MIST identifies sensitive parameters, then applies strong gradient dropout to regularize the update path, enabling the PTM to better align with task-specific distributions while maintaining generalizable representations.
Extensive experiments demonstrate that MIST can be seamlessly integrated into various freeze-based CL frameworks, consistently boosting performance across diverse datasets, especially under large distribution shifts.
The limitation of MIST lies in its reliance on efficient approximation of the Fisher matrix. When the data within a task exhibits significant distributional variation, this approximation may become inaccurate, potentially compromising the effectiveness of MIST. In the future, we plan to explore more robust Fisher estimation techniques that can adapt to intra-task variation.

\section*{Acknowledgment}

This paper was supported by the National Science Foundation of China (No. 62406323), China Postdoctoral Science Foundation (No. 2024M753496), and Postdoctoral Fellowship Program of CPSF (No. GZC20232993).

\bibliography{aaai2026}
\clearpage
\onecolumn
\appendix
\centerline{\Large\bfseries Appendices}
\begingroup
\fontsize{12pt}{14pt}\selectfont

\section{Gradient of Mutual Information} 
Mutual information gradients help pre-trained models acquire new knowledge while resisting the collapse of pre-trained knowledge. Below, we provide a detailed mathematical derivation to support this claim.

We begin with the standard definition of mutual information between two random variables \(X\) and \(Y\):
\begin{equation}
    I(X; Y) = \sum_{x,y} p(x, y) \log \frac{p(x, y)}{p(x)p(y)},
    \label{eq:mi_def}
\end{equation}
We now compute the gradient of \(I(X; Y)\) with respect to model parameters \(\theta\). Applying the chain rule, we obtain:
\begin{equation}
\begin{aligned}
\frac{\partial I(X; Y)}{\partial \theta}
&= \sum_{x,y} \frac{\partial p(x, y)}{\partial \theta} \log \frac{p(x, y)}{p(x)p(y)}
+ \sum_{x,y} p(x, y) \cdot \frac{\partial}{\partial \theta} \log \frac{p(x, y)}{p(x)p(y)} \\
&= \sum_{x,y} \frac{\partial p(x, y)}{\partial \theta} \log \frac{p(x, y)}{p(x)p(y)}\\
&+ \sum_{x,y} p(x, y) \left(
    \frac{1}{p(x, y)} \cdot \frac{\partial p(x, y)}{\partial \theta}
    - \frac{1}{p(x)} \cdot \frac{\partial p(x)}{\partial \theta}
    - \frac{1}{p(y)} \cdot \frac{\partial p(y)}{\partial \theta}
\right)
\end{aligned}
\label{eq:mi_grad_full}
\end{equation}
Note that:
\begin{equation}
\begin{aligned}
\sum_{x,y} p(x, y) \cdot \frac{1}{p(x)} \cdot \frac{\partial p(x)}{\partial \theta}
&= \sum_x \left( \frac{\partial p(x)}{\partial \theta} \cdot \sum_y \frac{p(x, y)}{p(x)} \right) \\
&= \sum_x \frac{\partial p(x)}{\partial \theta} \cdot \underbrace{\sum_y p(y \mid x)}_{=1}
= \sum_x \frac{\partial p(x)}{\partial \theta}.
\end{aligned}
\end{equation}
Similarly,
\begin{equation}
\sum_{x,y} p(x, y) \cdot \frac{1}{p(y)} \cdot \frac{\partial p(y)}{\partial \theta}
= \sum_y \frac{\partial p(y)}{\partial \theta} \cdot \sum_x \frac{p(x, y)}{p(y)}
= \sum_y \frac{\partial p(y)}{\partial \theta}.
\end{equation}
Because  \(\sum_x p(x) = \sum_y p(y) = 1\), their total derivatives must vanish:
\begin{equation}
\sum_x \frac{\partial p(x)}{\partial \theta} = 0, \quad
\sum_y \frac{\partial p(y)}{\partial \theta} = 0.
\end{equation}
The last two terms in Eq.~\ref{eq:mi_grad_full} cancel out, and the gradient simplifies to:
\begin{equation}
\frac{\partial I(X; Y)}{\partial \theta}
= \sum_{x,y} \frac{\partial p(x, y)}{\partial \theta} \log \frac{p(x, y)}{p(x)p(y)}.
\label{eq:mi_grad_final}
\end{equation}
By leveraging the normalization conditions of marginal distributions, the derivation shows how MI gradients inherently avoid the destabilizing term $\partial p(x; \theta)/\partial \theta$, which is present in CE-based optimization. This theoretical insight forms the foundation for our proposed MI-guided sparse tuning strategy, where we explicitly utilize MI gradients to identify stable and informative parameter directions for task-specific adaptation.

\section{Theoretical Justification of Batch-wise Gradient Accumulation.}
Let $g(x) = \frac{\partial \mathcal{L}_{\mathrm{MI}}(x)}{\partial \theta^i}$ denote the MI-based gradient with respect to parameter $\theta^i$. The exact gradient over the entire dataset $\mathcal{D}_t$ is:
\begin{equation}
    F_{\mathrm{MI}} = \left( \mathbb{E}_{x \sim \mathcal{D}_t} [g(x)] \right)^2.
\end{equation}
In practice, we approximate this expectation by averaging over $N$ mini-batches:
\begin{equation}
    F'_{\mathrm{MI}} = \left( \frac{1}{N} \sum_{j=1}^{N} g(x_j) \right)^2.
\end{equation}
According to the law of large numbers, if the mini-batches are drawn i.i.d. from $\mathcal{D}_t$ and the variance of $g(x)$ is sufficiently small, then:
\begin{equation}
\begin{aligned}
\mathbb{E}[F'_{\mathrm{MI}}] &= \mathbb{E}\left[ \left( \frac{1}{N} \sum_{j=1}^{N} g(x_j) \right)^2 \right] \\
&= \frac{\mathrm{Var}[g(x)]}{N} + \left( \mathbb{E}[g(x)] \right)^2
\xrightarrow{\mathrm{Var}[g(x)] \to 0} F_{\mathrm{MI}},
\end{aligned}
\end{equation}
where $\mathrm{Var}[g(x)]$ is the variance of $g(x)$.
This justifies the use of accumulated gradients across multiple mini-batches to estimate the MI-based Fisher scores in practice.
Moreover, in CL scenarios, the samples within $\mathcal{D}_t$ are typically uniform, which results in the value of $\mathrm{Var}[g(x)]$ being small, further reinforcing the validity of the approximation.

\section{Common tuning strategies for PTMs}

\textbf{Fully fine-tuning on PTMs}: In fully fine-tuning,  the model gains high plasticity as every parameter can be adapted to new tasks. However, this also maximally exposes the model to feature distribution drift due to the accumulation of large \( \frac{\partial p(x; \theta)}{\partial \theta^i} \) gradients across all parameters. As a result, the pre-trained generalization structure erodes rapidly, leading to instability across sequential tasks.

\noindent
\textbf{Naive partial fine-tuning on PTMs}: Naive partial fine-tuning methods attempt to reduce interference by limiting the number of updated parameters. For example, randomly updating a fixed proportion of parameters can help mitigate perturbations to \( p(x; \theta) \). However, the lack of guidance may still result in significant disruption to the pre-trained representation.
Other selection strategies, such as choosing parameters with the highest \( \ell_2 \) norm or those with the largest gradient magnitudes, inherently favor parameters that exhibit strong gradient responses, which may correspond to large values of \( \frac{\partial p(x; \theta)}{\partial \theta^i} \). As a result, even with a limited update scope, these methods still pose a substantial risk to the generalization ability of PTMs.

\noindent
\textbf{Fisher-guided partial fine-tuning on PTMs}:
Fisher-guided tuning methods provide another line of work, where sparse parameter updates are driven by estimated sensitivity scores (e.g., Fisher values). Higher Fisher scores often reflect large contributions from both \( \frac{\partial p(x, y; \theta)}{\partial \theta^i} \) and \( \frac{\partial p(x; \theta)}{\partial \theta^i} \). This suggests that Fisher-selected parameters, while effective for fast adaptation, are also more likely to induce substantial perturbations to the pre-trained feature distribution. Ironically, the parameters considered most ``important'' under the Fisher criterion are often those that inflict the greatest harm on generalization.

\section{Limitations of Batch-level MI Estimation and the Role of Sparse Tuning}
\begin{table}[t]
\centering
\caption{Final accuracy achieved under different batch sizes}

\label{tab:diff_batch}
\begin{tabular}{cccc}
\toprule
 batch size & 4  & 32  & 64 \\
\midrule
Final accuracy (\%)& 42.0  & 43.1  & 43.5 \\
\bottomrule
\end{tabular}
\end{table}
Although the MI loss provides a promising direction for preserving the pretrained representation, it still exhibits several critical limitations in practice:

\textbf{MI can only be estimated at the batch level.}  
The supervised InfoNCE loss defined in Eq.~(8) estimates MI $I(X; Y)$ using only mini-batch samples, limiting its representation of the true data distribution. Prior studies~\cite{guo2022ocm, oord2018representation} and our theoretical result in Eq.~\eqref{eq:mi_grad_final} indicate that more diverse batches improve MI estimation quality, as empirically confirmed in Table~\ref{tab:diff_batch}, where increasing the batch size from 4 to 64 raises final accuracy from 42.0\% to 43.5\%. Nevertheless, batch-wise computation inherently involves the term $\partial p(x; \theta)/\partial \theta^i$, causing unavoidable perturbations to the pretrained feature structure.

\textbf{Implicit disturbance arises from modeling $p(x, y; \theta)$.}  
Although the MI objective does not directly modify the marginal input distribution $p(x; \theta)$, it optimizes the joint distribution $p(x, y; \theta)$ to encourage discriminative representations. This can implicitly shift the geometry of the feature space learned by the PTM, leading to misalignment with the original pretrained structure and reduced generalization.

\textbf{Sparse tuning helps mitigate the above limitations.}  
To address these issues, we adopt MI-guided parameter selection and gradient dropout as regularization strategies. Specifically, we select only the top-$k\%$ most MI-sensitive parameters and randomly drop $d\%$ of them in each batch, resulting in only $0.5\%$ of parameters being updated. This strong sparsity reduces the risk of feature drift, mitigates overfitting to local updates, and preserves structural integrity during task adaptation. Hence, sparse tuning not only improves efficiency but also plays a crucial role in stabilizing the adaptation process, making it an essential complement to batch-level MI optimization. Figure~\ref{fig:mist_improvements} presents the final accuracy comparisons on ImageNet-R and Cars196. Across all baseline methods, incorporating MIST consistently yields notable accuracy enhancements. This highlights the general effectiveness of our proposed MIST strategy in improving incremental learning performance.

\begin{figure}[t]
  \subfigure[ImageNet-R dataset]{
    \vspace{-10px}
    \includegraphics[width=0.49\textwidth]{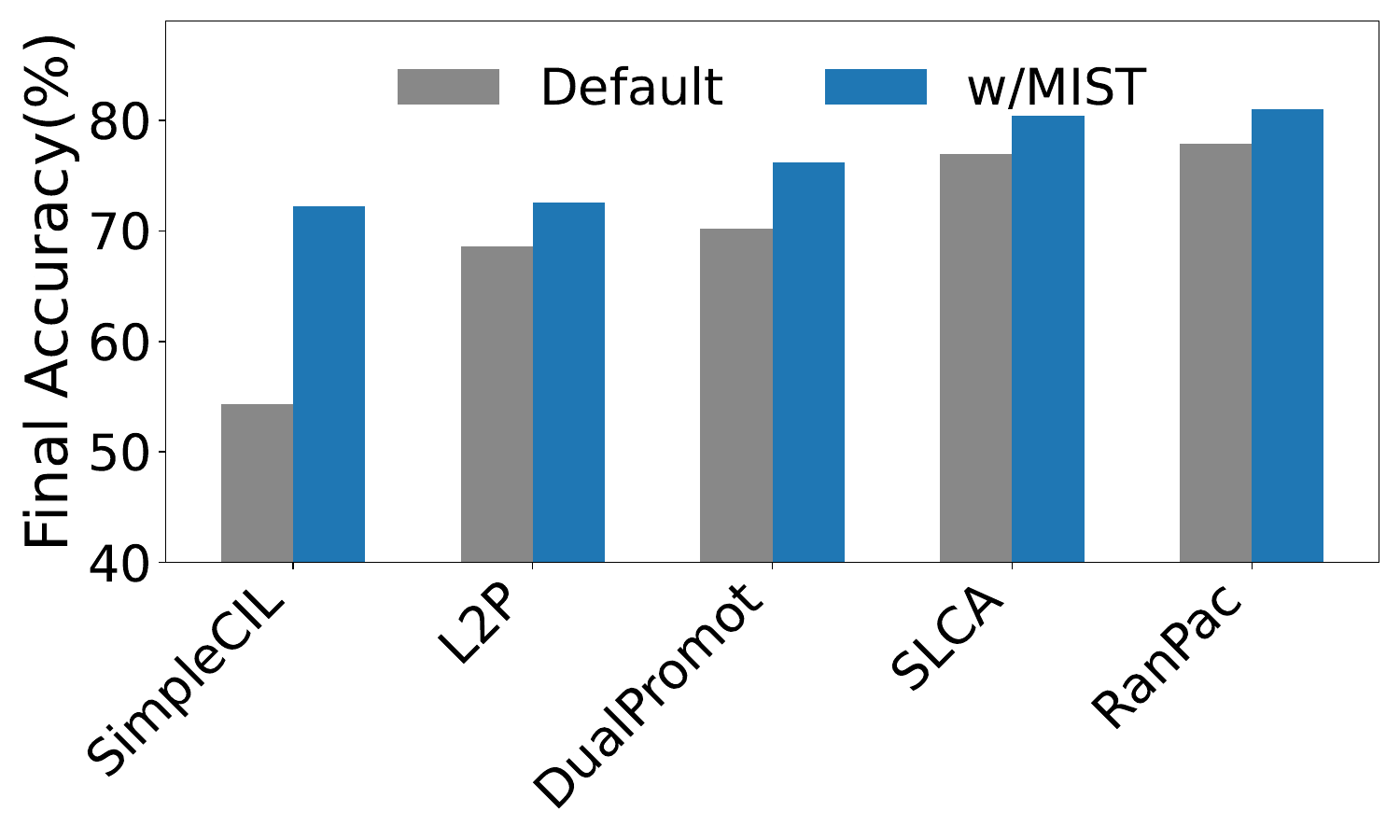}
  }
  \hfill 	
  \subfigure[Cars dataset]{
    \includegraphics[width=0.49\textwidth]{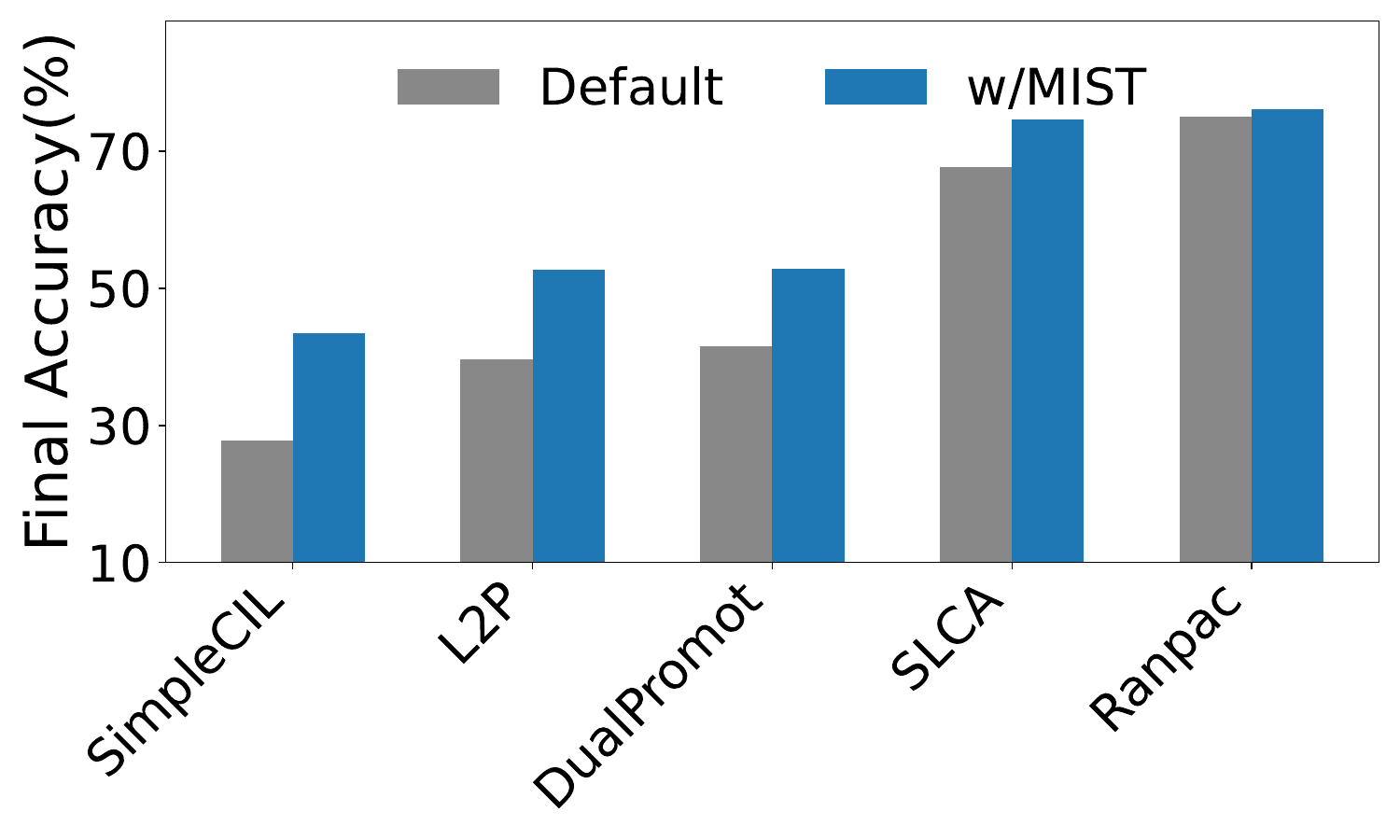}
  }
  \vspace{-10px}
\caption{
Performance comparison of incremental learning methods with and without the proposed MIST. The inclusion of MIST consistently improves final accuracy across all evaluated methods and datasets.
}
\label{fig:mist_improvements}
\end{figure}

\begin{table}[t]
  \centering
  \begin{minipage}[t]{0.48\textwidth}
    \centering
    \caption{Different select rates $k\%$.}
    \label{tab:select_rate}
    \begin{tabular}{ccccc}
      \toprule
      \multirow{2}{*}{$k\%$} 
        & \multicolumn{2}{c}{\textbf{ImageNet-R}}
        & \multicolumn{2}{c}{\textbf{Cars196}} \\ 
      & $\bar{A}$ & $A_T$ & $\bar{A}$ & $A_T$ \\ 
      \midrule
      20  & 84.1 & 80.0 & 82.1 & 74.8 \\ 
      10   & \textbf{85.0} & 80.8 & 82.7 & 76.0\\
      5    & 84.7 & \textbf{81.0} &  \textbf{83.1} &  \textbf{76.4}\\
      1    & 84.6 & 80.2 & 82.6 & 76.0\\
      0.1  & 84.4 & 79.6 & 81.5 & 74.1\\
      \bottomrule
    \end{tabular}
  \end{minipage}\hfill
  \begin{minipage}[t]{0.48\textwidth}
    \centering
    \caption{Different drop rates $d\%$.}
    \label{tab:drop_rate}
    \begin{tabular}{ccccc}
      \toprule
      \multirow{2}{*}{$d\%$} 
        & \multicolumn{2}{c}{\textbf{ImageNet-R}}
        & \multicolumn{2}{c}{\textbf{Cars196}} \\ 
      & $\bar{A}$ & $A_T$ & $\bar{A}$ & $A_T$ \\ 
      \midrule
      0    & 82.2 & 76.7 & 78.0 & 69.1 \\ 
      50   & 83.3 & 78.8 & 81.2 & 73.7\\
      80   & 84.7 & 80.8 & \textbf{83.3} & \textbf{76.5}\\
      90   & \textbf{84.7} & \textbf{81.0} & 83.1 & 76.4\\
      99   & 80.8 & 76.9 & 76.4 & 67.1\\
      \bottomrule
    \end{tabular}
  \end{minipage}
\end{table}

\section{Effect of selection rate and dropout rate.}
We conduct a hyperparameter study to explore how the parameter selection rate $k\%$ and the gradient dropout rate $d\%$ affect the performance of MIST, as reported in Table~\ref{tab:select_rate} and Table~\ref{tab:drop_rate}, respectively.
We observe that high sparsity generally yields better performance. For instance, selecting only 5\% of parameters per task achieves $A_T = 81.0\%$ on ImageNet-R and $76.4\%$ on Cars196, which outperforms full fine-tuning. This validates that MIST is able to effectively adapt to new tasks while preserving the pre-trained structure by updating only a small subset of critical parameters.
The performance of $k=5\%$ is very close to $k=10\%$, with only marginal differences. Given that fewer parameters are involved and the computational cost is lower, we adopt $k=5\%$ as the default in practice.
Table~\ref{tab:drop_rate} shows that increasing the gradient dropout rate significantly improves performance, especially from $d=0\%$ to $d=90\%$. This confirms that dropout acts as an effective regularizer, helping to suppress local gradient bias and mitigate overfitting to static parameter importance scores. The best performance is observed when $d = 90\%$, and we use this setting as the default for all experiments.

\begin{figure}[t]
  \subfigure[ImageNet-A]{
    \includegraphics[width=0.22\textwidth]{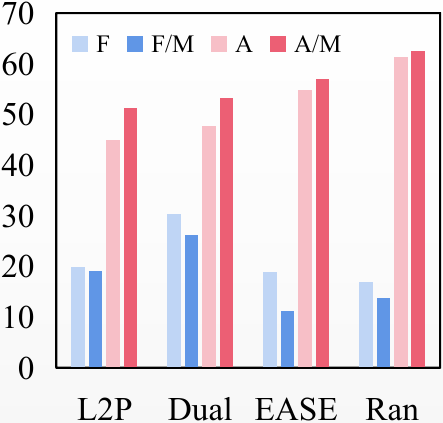}
  }
  \hfill 	
  \subfigure[ImageNet-R]{
    \includegraphics[width=0.22\textwidth]{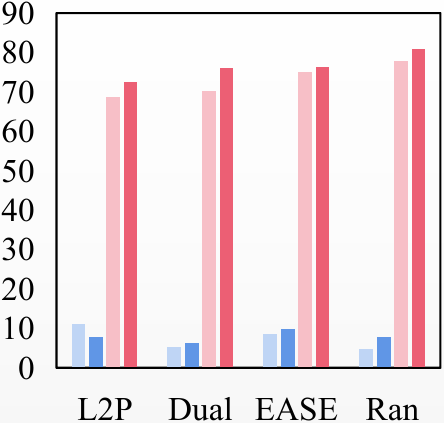}
  }
  \hfill 	
  \subfigure[Cifar-100]{
    \includegraphics[width=0.22\textwidth]{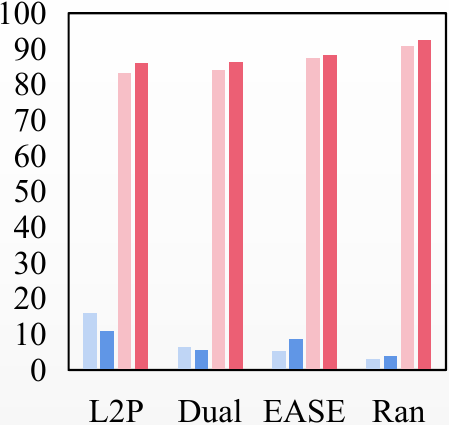}
  }
    \hfill 	
  \subfigure[Cub]{
    \includegraphics[width=0.22\textwidth]{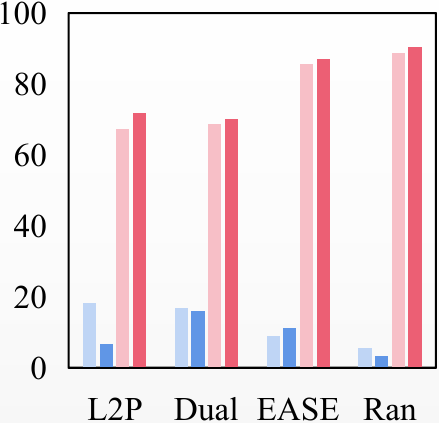}
  }
\caption{Comparison of average forgetting F and accuracy A across datasets. F/M and A/M represent results with MIST integration. ``Dual'' denotes DualPrompt; ``Ran'' denotes RanPac.}
\label{fig:forgetting}
\end{figure}

\section{Impact of MIST on Forgetting and Learning Performance}

To further compare the performance of different methods after integrating MIST, we report the average forgetting~\cite{kang2022forget} in Figure~\ref{fig:forgetting}. Average Forgetting measures the degree of knowledge retention loss for each task after training on the final task, and is defined as:
\begin{equation}
    \text{Average Forgetting} = \frac{1}{T-1} \sum_{j=1}^{T-1} d_{T,j},
\end{equation}
where \( d_{T,j} \) denotes the forgetting on task \( j \) after training up to task \( T \), computed as:
\begin{equation}
    d_{i,j} = \max_{k \in \{1,\ldots,i-1\}} a_{k,j} - a_{i,j},
\end{equation}
with \( a_{i,j} \) representing the test accuracy on task \( j \) after training on task \( i \). This metric captures the worst-case drop in performance on previous tasks over the training trajectory.

As shown in the results, while incorporating MIST generally reduces the forgetting rate across most methods and datasets, there are instances where forgetting increases. This can be attributed to MIST’s core mechanism—enhancing the efficiency of learning new knowledge.
In some cases, a lower forgetting rate without MIST does not imply better retention of prior knowledge. Instead, it may result from poor learning of new tasks, where limited improvement on recent tasks leads to a small performance gap between old and new tasks, thus producing an artificially low forgetting rate. Conversely, with MIST, some methods exhibit higher forgetting because MIST significantly boosts performance on new tasks. This amplifies the gap between old tasks (with stable or slightly degraded performance) and new tasks (with much improved accuracy), leading to a higher measured forgetting rate.
Importantly, even in these cases, the overall average accuracy across all tasks improves with MIST. This reflects a more favorable balance between stability (retaining old knowledge) and plasticity (learning new knowledge). In incremental learning, a moderate increase in forgetting is acceptable—and often expected—when overall learning improves. It signals better adaptation to new tasks, which is the core goal: to acquire new knowledge effectively while maintaining reasonable retention of previous knowledge, rather than minimizing forgetting at the cost of stagnating new task performance.

\section{Experiments on Domain-Incremental Learning}

To further validate the general applicability of our method, we also conduct experiments on domain-incremental learning using the DomainNet dataset~\cite{domainnet}, which is a large-scale benchmark dataset containing images from 345 categories across six diverse visual domains. Table~\ref{tab:dil} compares the proposed SimpleCIL+MIST method with several baseline approaches, including L2P, Adam, and SimpleCIL. As shown, SimpleCIL+MIST achieves the highest final accuracy (53.5\%), indicating its effectiveness in DIL scenarios as well.

\begin{table}[t]
\centering
\caption{Comparison of different methods for domain-incremental learning evaluated on the DomainNet dataset. The proposed SimpleCIL+MIST achieves the highest final accuracy.}
\label{tab:dil}
\begin{tabular}{lcccc}
\toprule
Method & L2P  & Adam  & SimpleCIL& SimpleCIL+MIST \\
\midrule
Final accuracy (\%)& 40.2  & 50.3  & 49.5& \textbf{53.5} \\
\bottomrule
\end{tabular}
\end{table}

\end{document}